\documentclass[runningheads]{llncs}
\usepackage{marvosym}
\usepackage{tablefootnote}
\usepackage{amssymb}
\usepackage{mathrsfs}
\usepackage{amsmath}
\usepackage{multirow}
\usepackage{makecell}
\usepackage{graphicx}
\usepackage[colorlinks=true,
            linkcolor=blue,
            anchorcolor=blue,
            citecolor=blue]{hyperref}
\begin{document}
\title{Memoryless Multimodal Anomaly Detection via Student-Teacher Network and Signed Distance Learning \thanks{Corresponding author: Zhongbin Sun. Supported by the Fundamental Research Funds for the Central Universities under Grant No. 2021QN1075.}}

\titlerunning{Memoryless Multimodal Anomaly Detection via S-T Network and SDL}
%
\author{Zhongbin Sun\inst{1, 2} $^{\href{mailto: zhongbin@cumt.edu.cn}{\textrm{(\Letter)}}}$ \and
Xiaolong Li\inst{2} $^{\href{mailto: lixiaoloong@cumt.edu.cn}{\textrm{(\Letter)}}}$ \and
Yiran Li\inst{3} $^{\href{mailto: 07220333@cumt.edu.cn}{\textrm{(\Letter)}}}$\and
Yue Ma\inst{2} $^{\href{mailto: 08222214@cumt.edu.cn}{\textrm{(\Letter)}}}$}

\authorrunning{Z. Sun et al.}

\institute{Mine Digitization Engineering Research Center of the Ministry of Education, Xuzhou, Jiangsu, 221116, China \and School of Computer Science and Technology, China University of Mining and Technology, Xuzhou, Jiangsu, 221116, China\\
\and Sun Yueqi Honors College, China University of Mining and Technology, Xuzhou, Jiangsu, 221116, China\\
}

\maketitle              
\begin{abstract}
Unsupervised anomaly detection is a challenging computer vision task, in which 2D-based anomaly detection methods have been extensively studied. However, multimodal anomaly detection based on RGB images and 3D point clouds requires further investigation. The existing methods are mainly inspired by memory bank based methods commonly used in 2D-based anomaly detection, which may cost extra memory for storing mutimodal features. In present study, a novel memoryless method MDSS is proposed for multimodal anomaly detection, which employs a light-weighted student-teacher network and a signed distance function to learn from RGB images and 3D point clouds respectively, and complements the anomaly information from the two modalities. Specifically, a student-teacher network is trained with normal RGB images and masks generated from point clouds by a dynamic loss, and the anomaly score map could be obtained from the discrepancy between the output of student and teacher. Furthermore, the signed distance function learns from normal point clouds to predict the signed distances between points and surface, and the obtained signed distances are used to generate anomaly score map. Subsequently, the anomaly score maps are aligned to generate the final anomaly score map for detection. The experimental results indicate that MDSS is comparable but more stable than the SOTA memory bank based method Shape-guided, and furthermore performs better than other baseline methods.

\end{abstract}
\section{Introduction}
Visual anomaly detection aim to detect abnormal objects from visual information, which is widely used in industrial and medical imaging fields~\cite{ruff2021unifying}. In practical application scenarios, due to the low proportion of abnormal areas, unknown abnormal patterns and expensive annotation costs, it is difficult to obtain high-quality labeled datasets. Therefore, unsupervised anomaly detection has been the subject of researcher interest, in which previous research has mainly focused on 2D anomaly detection with RGB images~\cite{liu2024deep}. 

With the proposal of the MVTec 3D-AD dataset~\cite{bergmann2021mvtec3d} in 2022, researchers have begun to study the feasibility of combining 3D point clouds with RGB images for multimodal anomaly detection~\cite{Bergmann20233DST,chu2023shape,costanzino2023multimodal,horwitz2023back,rudolph2023asymmetric,wang2023multimodal}. The key to unsupervised multimodal anomaly detection lies in how to integrate information from two modalities to distinguish normal and abnormal samples. We categorize the existing methods into two classes: (1) student-teacher network based methods, (2) memory bank based methods.

\begin{figure}[!h]
    \centering
    \includegraphics[width=\textwidth]{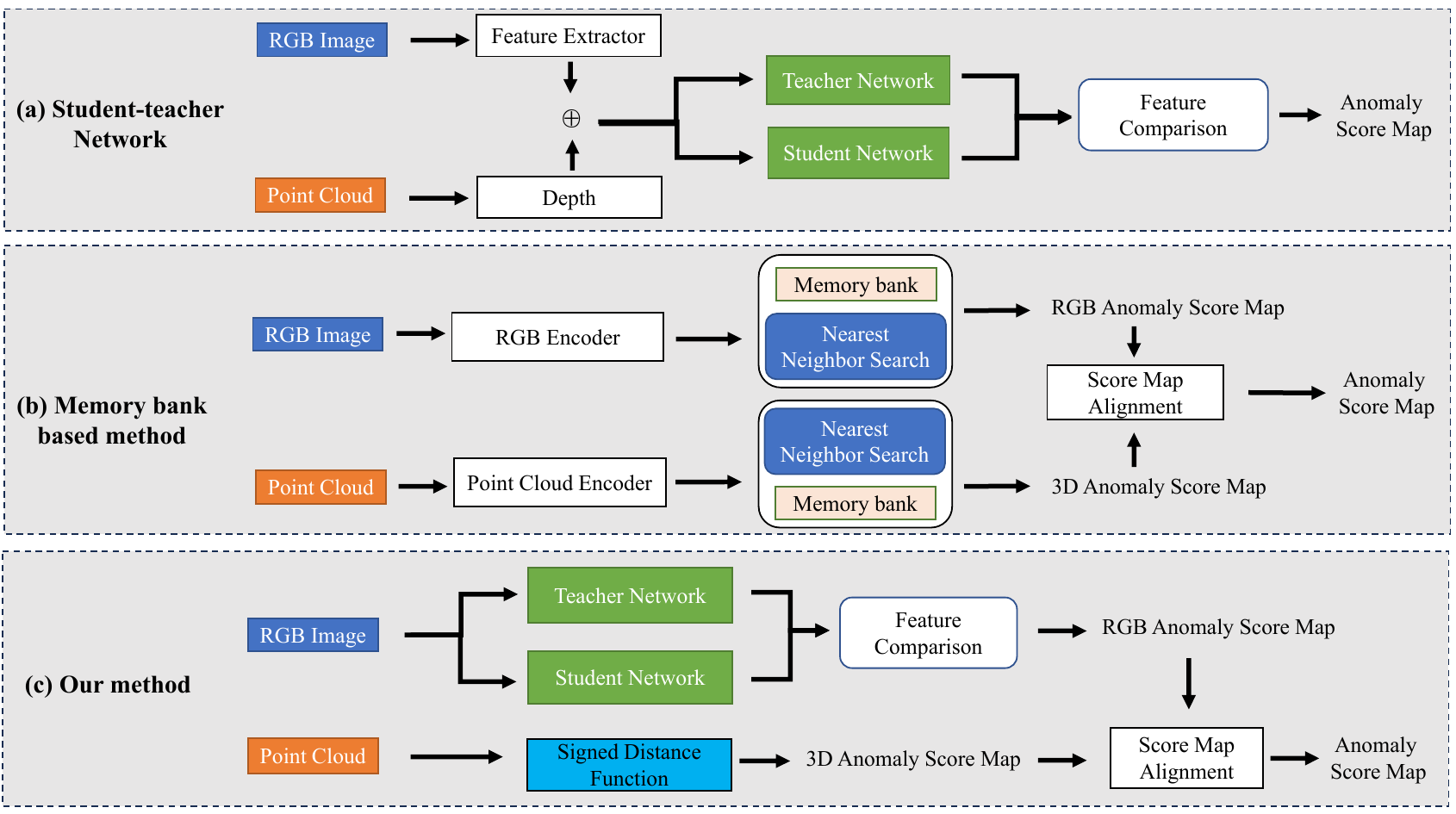}
    \caption{Comparison of the structure of our method with that of other methods.}
    \label{fig:frame comparison}
\end{figure}

In methods based on student-teacher network,
the RGB features and depth from point clouds are concatenated and input into the student-teacher network for anomaly detection~\cite{rudolph2023asymmetric}, as shown in Fig.~\ref{fig:frame comparison}(a). Particularly, the output difference between the teacher and the student can be regarded as anomaly scores for indicating the possibility of a sample to be abnormal. However, only using depth discards the 3D information in the coordinate of $x$ and $y$ and direct concatenation may cause disturbance between the features of different modalities, which harm the detection performance.

In methods based on memory bank, the features of normal samples of different modalities are stored ~\cite{chu2023shape,horwitz2023back,wang2023multimodal}, as shown in Fig.~\ref{fig:frame comparison}(b). For the inference stage, the new samples are input into the same feature extractor, and corresponding output features are compared with the stored features of normal samples  to generate anomaly scores. However, these memory bank based methods consume extra memory and have high hardware requirements, which may be difficult to apply in some real application scenarios.

To address the aforementioned issues, a \textbf{M}emoryless multimodal anomaly \textbf{D}etection method by combining \textbf{S}tudent-teacher network and \textbf{S}igned distance learning (\textbf{MDSS}) is proposed, as shown in Fig.~\ref{fig:frame comparison}(c). In MDSS, for the purpose of reducing disturbance between modalities, the student-teacher network is used to process RGB images and the signed distance function is employed to process 3D point clouds respectively. 

To be specific, in student-teacher network, RGB images with masks generated from 3D point clouds are employed to obtain the RGB anomaly score map. Moreover, existing research~\cite{batzner2024efficientad} shows that excessive training images may lead to homogenization of the student and teacher, while insufficient training images may hinder the student from learning the features of normal samples, therefore MDSS employs a dynamic learning factor in loss to train the student-teacher network for improving the anomaly detection performance. In addition, to the best of our knowledge, we are the first to propose the direct utilization of signed distance function for unsupervised 3D anomaly detection. For signed distance learning in MDSS, the signed distance function is employed for surface reconstruction from normal point clouds and outputs the distances from the points to the surface. We assume that the distance between abnormal point and the surface is larger than that of normal point, thus the distance can be used to measure the possibility of a point to be anomaly and corresponding 3D anomaly score map is obtained. Finally, a statistical approach is employed to align the RGB anomaly score map and 3D anomaly score map. Then the aligned score map which combines the anomaly information from both RGB and 3D is used for image-level and pixel-level anomaly detection.

In the experimental study, the popular MVTec 3D-AD dataset~\cite{bergmann2021mvtec3d} is used and several representative multimodal anomaly detection methods are selected as the baseline methods. The experimental results show that  MDSS is comparable but more stable than the SOTA memory bank based method Shape-guided, and furthermore performs better than other baseline methods. In addition, we also conduct an ablation study to demonstrate the effectiveness of combining the student-teacher network and signed distance learning.

The main contributions of our work can be concluded as the follows:
\begin{itemize}
    \item[$\bullet$] A memoryless multimodal anomaly detection method is proposed by directly utilizing the output of student-teacher network and signed distance function.
    \item[$\bullet$] We are the first to use signed distance for 3D unsupervised anomaly detection, reducing the usage of memory for inference.
    \item[$\bullet$] The proposed method achieves the SOTA image-level multimodal anomaly detection performance in terms of I-AUROC.
\end{itemize}

\section{Related Work}

\subsection{Student-teacher network}

Student-teacher network is originally an approach for knowledge distillation~\cite{hinton2015distilling}. Bergmann et al.~\cite{bergmann2020uninformed} first applies the student-teacher network to 2D anomaly detection. They assume significant regression errors between student and teacher networks in anomaly representation as well as substantial uncertainty in anomaly representation among multiple student networks can be used for anomaly detection. The STPFM~\cite{Wang2021STFPM} introduces the integration of a multi-scale feature matching strategy into the student-teacher framework which enables the detection of anomalies of diverse sizes. Deng et al.~\cite{Deng2022ReverseDistill} present a novel student-teacher network comprising a teacher encoder and a student decoder, along with the introduction of a straightforward yet powerful "reverse distillation" paradigm. Their student network accepts the one-class embedding from the teacher model as input and aims to reconstruct the teacher's multi-scale representations. EfficientAD~\cite{batzner2024efficientad} utilizes an autoencoder and a lightweight student-teacher network trained by an asymmetric loss function to obtain a combined anomaly map complementing both global and local anomaly information to improve detection accuracy, while also improving computational efficiency.

In the field of multimodal anomaly detection, Bergmann et al.~\cite{Bergmann20233DST} construct an expressive teacher network that extracts dense local geometric descriptors and regression errors between the teacher and the student are utilized to achieve reliable localization of anomalous structures. Rudolph et al.~\cite{rudolph2023asymmetric} proposes asymmetric student-teacher networks (AST). To be specific, They train a normalizing flow for density estimation as the teacher and a conventional feed-forward network as the student to induce significant distances for anomalies. In addition, the RGB features and depth from point clouds are concatenated to train the student-teacher network for multimodal anomaly detection.

\subsection{Memory bank based methods}
Memory bank based methods initially originate from 2D anomaly detection.
Cohen et al.~\cite{cohen2020sub,Reiss2021PANDA} propose to store the deep pretrained features and use the K nearest neighbors of features extracted from a new sample to conduct both anomaly detection and localization. This approach later becomes the foundation of many memory bank based methods. Many researchers begin to investigate such methods in 2D anomaly detection. The most representative method is PatchCore~\cite{roth2022towards}. PatchCore realizes a maximally downsampled representative memory bank via greedy coreset subsampling, which comprises locally aware, nominal patch-level feature representations extracted from ImageNet pretrained networks.

In multimodal anomaly detection, researchers mainly migrate the memory bank based methods from 2D anomaly detection. BTF~\cite{horwitz2023back} combines handcrafted 3D representations (FPFH) with a deep, color-based method (PatchCore), outperforming the baseline provided by the author of MVTec 3D-AD by a large margin and justifying that there are complimentary benefits from using both 3D and color modalities. M3DM~\cite{wang2023multimodal} employs different backbones for RGB images and point clouds respectively, and designs an unsupervised feature fusion with patch-wise contrastive learning to encourage the interaction of different modal features which are stored in multiple memory banks for final detection. However, with the memory banks, the memory cost of M3DM in inference can be 6.52GB and the FPS is only 0.51 which is not unacceptable in some real application scenarios~\cite{costanzino2023multimodal}.
Shape-guided~\cite{chu2023shape} uses the two experts (ResNet for RGB images and signed distance function for point clouds) to build the dual memory banks from the anomaly free training samples and performs shape-guided inference. 

In conclusion, the drawback of current methods based on student-teacher network is that the RGB features and point clouds are directly concatenated, which may result in interference between modalities. Furthermore, using only depth leads to the loss of some 3D information in the point clouds. Moreover, memory bank based methods suffer from the disadvantage of excessive memory consumption. In addition to the above two types of methods, Costanzino et al. recently propose a novel method CFM~\cite{costanzino2023multimodal}, in which a novel light and fast framework is introduced for learning to map features from one modality to the other on normal samples. During inference, anomalies are detected by pinpointing inconsistencies between observed and mapped features. CFM achieves faster inference and occupies less memory than memory bank based methods, which will be selected as one of the baseline methods in present study.
 
\section{Method}
\begin{figure}[!htbp]
    \centering
    \includegraphics[width=\textwidth]{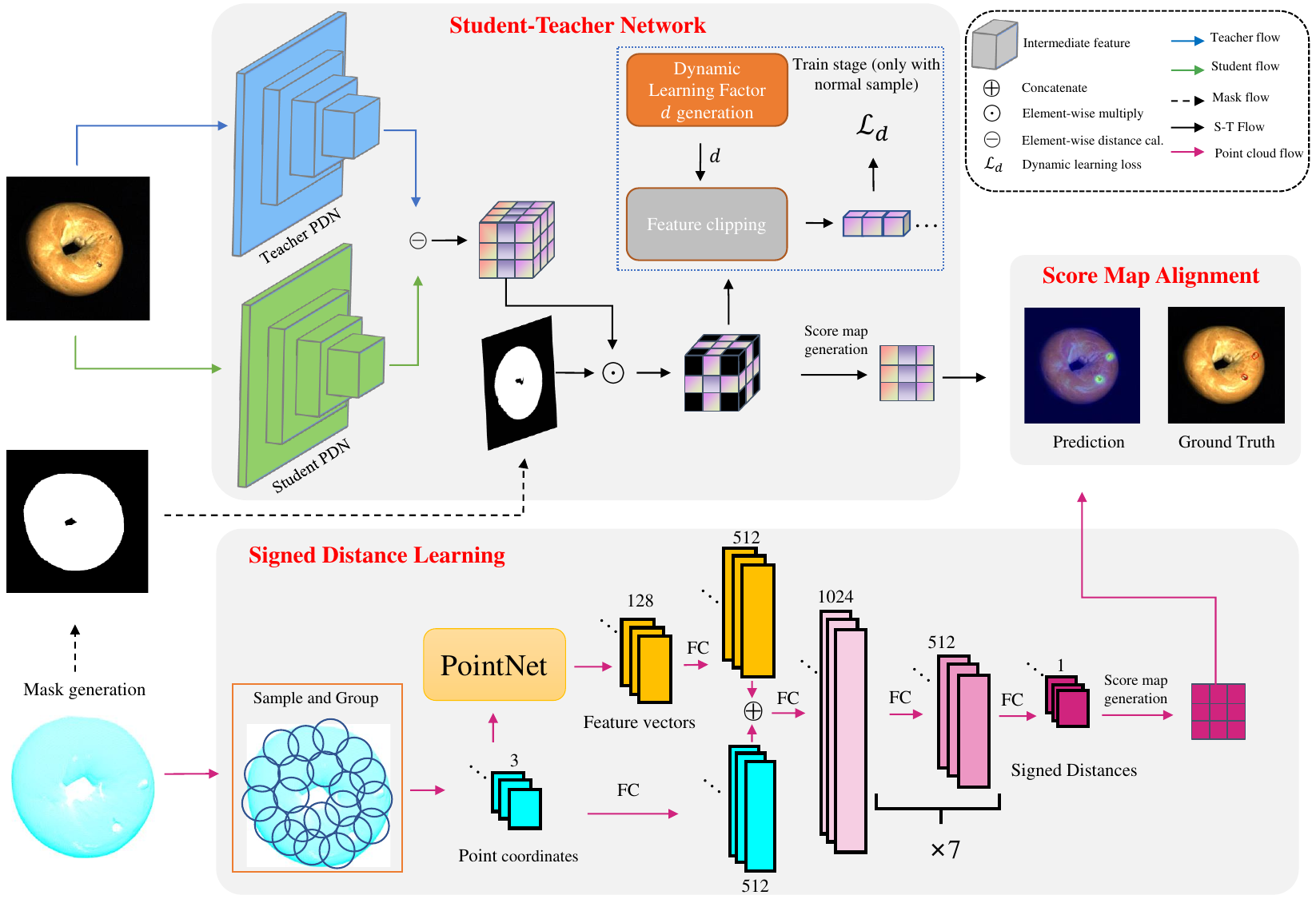}
    \caption{The framework of our method MDSS.}
    \label{fig:model structure}
\end{figure}

In our memoryless multimodal anomaly detection method MDSS, three modules are included, respectively student-teacher network, signed distance learning and score map alignment. Fig.~\ref{fig:model structure} provides the detailed framework of the proposed method MDSS.

To be specific, the student-teacher network is trained with normal RGB images and masks generated from point clouds by a dynamic loss, and the anomaly score map could be obtained from the discrepancy between the output of student network and teacher network. Moreover, in signed distance learning, the signed distance function learns from normal point clouds to predict the signed distances between points and surface, and the obtained signed distances are used to generate anomaly score map. Subsequently, the two previously obtained anomaly score maps are aligned to generate the final anomaly score map for detection.

In the following, Section~\ref{sec:student-teacher network} provides the detail of our student-teacher network. Section~\ref{sec:signed distance learning} presents the process of signed distance learning. Finally, anomaly score map alignment will be introduced in Section~\ref{sec:score map alignment}.

\subsection{Student-Teacher Network}\label{sec:student-teacher network}

Student-teacher network has been widely used in 2D anomaly detection. In MDSS, the structure of the student network is same as the teacher. Particularly, a light-weighted student-teacher network PDN (Patch Description Network)~\cite{batzner2024efficientad} is selected to detect the anomaly in RGB images, which contains four convolution layers leading to a very low overall latency and employs a limited receptive field that can localize context-related anomaly.

In the proposed method MDSS, a training image $I$ is applied to the Teacher $T$ and Student $S$, and corresponding features $T(I)\in \mathbb{R}^{C\times H\times W}$ and $S(I)\in \mathbb{R}^{C \times H\times W}$ are obtained. The square difference of each tuple $(c,w,h)$ is computed as $D_{c, w, h}=(T(I)_{c,w,h}-S(I)_{c,w,h})^2$. Moreover, as the object is presented in a 3D perspective with static background, it is straightforward and reasonable to remove the irrelevant background, which is the case for almost all real applications. Therefore, a binary mask $M$ is generated from 3D point clouds for extracting the foreground of the object. Then the element-wise product $\odot$ is used to $D$ and $M$ for setting the output elements belonging to the background as zero.

Moreover, excessive training may cause the student to mimic the output of teacher beyond normal samples, thereby hindering the detection accuracy~\cite{batzner2024efficientad}, PDN employs a hard feature loss for training, which uses the output elements with the highest loss for back propagation to encourage the student to focus on emulating the most underfitting regions. However, they ignore that the proportion of these regions will change dynamically during the training process. Furthermore, the fixed proportion may not be appropriate for different datasets. Therefore a dynamic learning factor $d$ is proposed in MDSS to solve this problem. 

Specifically, the dynamic learning factor $d$ varies between 0.99 and 0.999 in a cosine annealing~\cite{loshchilov2016sgdr} way. Given a dynamic learning factor $d$,  the $d$-quantile of the elements in $D\odot M$ is computed and the elements larger than $d$-quantile are averaged as our dynamic loss, referred as $\mathcal{L}_{d}$ and calculated in Equation~\ref{eq1:loss}. 

\begin{equation} \label{eq1:loss}
\mathcal{L}_{d} =\frac{1}{n} \sum_{i=1}^{n} DM_i
\end{equation}
where $DM_{i} \in \{D^{'} \mid D^{'} \in D \odot M, D^{'}>d\text{-quantile}\}$, $n$ represents the number of elements in the set.

In inference stage, the trained student-teacher network is employed to generate corresponding anomaly score map for a new image. Particularly, the distance between the output of the teacher and the student is firstly calculated and averaged along the channel dimension to obtained the anomaly score map. Then the score map is resized to the size of input image with bilinear interpolation. Each pixel value in the map represents the likelihood of an anomaly pixel and the maximum value of the score map is regarded as the anomaly score on image-level.

\subsection{Signed Distance Learning}\label{sec:signed distance learning}

Chu et al.~\cite{chu2023shape} introduce the signed distance function (SDF) for 3D anomaly detection and propose the method Shape-guided. In Shape-guided, SDF is applied to the point clouds and store the SDF features into memory bank for inference. SDF is a continuous function to output the distance of a point to the closest surface, in which the sign represents whether the point is inside or outside the watertight mesh, and the underlying surface boundary is implicitly represented by the zero-level set with the distance being zero. Due to spatial locality of the occurrence of anomaly, Shape-guided employs PointNet~\cite{qi2017pointnet} and Neural Implicit Function~\cite{ma2022surface} to obtain the local geometry representation and store them into the memory bank.

However, in our opinion, if the SDF is trained only with normal samples, the model will be learned to predict the distances of normal points to the implicit surface. Therefore in inference, the distance from abnormal point to surface is expected to be larger than that from normal point, and it can be used as the anomaly score. Therefore, MDSS directly uses SDF for 3D anomaly detection without the usage of memory bank. Particularly, MDSS trains the SDF model for each category respectively and the output signed distance is directly used to detect anomaly. Note that the PointNet in SDF is trained from scratch.

For inference stage in signed distance learning, a new point cloud sample is passed into the SDF model, and the signed distances between points and the surface are output. With corresponding 2D index, the signed distances of each point can be assigned to a pixel in anomaly score map. Furthermore, Gaussian blur is applied to the anomaly score map for improving the relevance between the anomaly point and its neighbours. Similar to~\ref{sec:student-teacher network},  the maximum value of the score map is regarded as the anomaly score on image-level.

\subsection{Score Map Alignment}\label{sec:score map alignment}
Our method MDSS is built upon the capability of these two models and their collaborative nature to more effectively tackle the challenge of multimodal anomaly detection. To be specific, the student-teacher network considers the RGB information to identify any appearance irregularities in the aspect of color, and the signed distance learning utilizes 3D information to probe possible anomalies in shape geometry. As mentioned in~\ref{sec:student-teacher network} and~\ref{sec:signed distance learning}, the two models  will respectively generate an anomaly score map. The product of the maximum of these two maps is used for image-level anomaly detection.

However, for pixel-level anomaly detection (also referred as anomaly segmentation), the anomaly scores in the two score maps should be firstly transformed into a similar scale due to their significant numerical gap. In MDSS, a statistical method is used to align them. Specifically, the validation set is employed to simulate the distribution of the two score maps in real scenarios. Then we compute the mean value and standard deviation of RGB and 3D anomaly score maps respectively. During inference stage, the RGB scores for a new sample will be aligned to its 3D scores with the previously obtained mean value and standard deviation, such that the mean $\pm$ 3$\times$standard deviation of RGB scores is aligned to the mean $\pm$ 3$\times$standard deviation of 3D scores. The pixel-wise maximum of two score maps is selected to construct the final anomaly score map for anomaly segmentation.

\section{Experiments}
\subsection{Dataset \& Evaluation Metrics}
The proposed method MDSS is validated for its effectiveness with the popular MVTec 3D-AD~\cite{bergmann2021mvtec3d} dataset in the experimental study. This dataset is the first publicly available multimodal anomaly detection dataset and comprises 10 categories, including natural objects and industrial components. Particularly, MVTec 3D-AD contains 2656 training, 294 validation, and 1197 test samples, where the test data are split into 249 normal samples and 948 abnormal samples. The abnormal test samples include about 4 to 5 different types of defects in each category. 

To be specific, each category is represented by both RGB images and high-resolution 3D point clouds. The 3D point clouds are obtained using structured light from an industrial sensor and store position information in 3-channel tensors (x, y, and z coordinates), while RGB information is recorded for each point. Since all samples in the dataset are viewed from the same angle, the RGB information for each sample can be stored into a single image. Furthermore, the labels and pixel ground truths are also provided to conduct image-level detection and pixel-level detection.

As is common for anomaly detection, we adopt the area under the receiver operator curve (AUROC) to evaluate the detection performance of our method on image-level(I-AUROC). For segmentation evaluation, the per-region overlap (PRO) metric~\cite{bergmann2019mvtec} is employed, which is defined as the average relative overlap of the binary prediction with each connected component of the ground truth. Similar to I-AUROC, the area under PRO curve (AUPRO) is computed to evaluate the pixel-level detection performance. Note that both metrics range from 0 to 1 and higher values indicate better performance.

\subsection{Implementation Detail}

For training our model, the hyper-parameters and corresponding values are provided in Table~\ref{tab:hyper-parameters of our model}. 

\begin{table}[!htbp]
    \caption{Hyper-parameters of the proposed method MDSS.}
    \centering
    \tabcolsep=0.15cm
    \resizebox{\textwidth}{!}{
    \begin{tabular}{l c c}
    \hline
     \textbf{Parameters} & \textbf{Student-Teacher Network} & \textbf{Signed Distance Learning}\\
    \hline
    learning rate & 0.001 & cosine annealing~\cite{loshchilov2016sgdr}\\ 
    batch size & 4 & 32 \\ 
    mask &  $\checkmark$ & - \\
    dynamic learning factor & cosine annealing~\cite{loshchilov2016sgdr} & - \\
    point of each patches& - &500\\
    \hline
    \end{tabular}}
    \label{tab:hyper-parameters of our model}
\end{table}

In the experiment, the background plane of the point clouds are removed according to the method in~\cite{horwitz2023back}. Then all of the point clouds are cut into different patches and each patch includes 500 points. Specifically, a set of points from the original points is sampled with farthest point sampling~\cite{qi2017pointnet++} and the $K$-nearest points to each of them are searched to construct a patch. In present experiment, we set default value of $K$ to 500. Note that the patches may overlap with each other and each point should belongs to at least one patch. 

Furthermore, inspired by~\cite{rudolph2023asymmetric}, the point clouds are used to generate corresponding mask $M$ for RGB images. Specifically, if there is a non-zero pixel in point clouds images, we set it to 1 at the same position in the mask for foreground, otherwise 0 for the background. In order to fill missing values, the foreground mask is dilated using a square structural element of size 8. Both of point clouds and RGB images are resized to $256 \times 256$.

\subsection{Experimental Results}
In the experiment, we compare MDSS with five different methods on 10 category of MVTec 3D-AD, including BTF~\cite{horwitz2023back}, AST~\cite{rudolph2023asymmetric}, M3DM~\cite{wang2023multimodal}, Shape-guided~\cite{chu2023shape}, and CFM~\cite{costanzino2023multimodal}. Table~\ref{tab:performance image-auroc} and  Table~\ref{tab:performance aupro} respectively provide the detection and segmentation performance in terms of I-AUROC and AUPRO. Note that the best performance is highlighted in bold. What's more, the AUPRO results of AST is not presented in their study and CFM cannot provide results with only RGB or 3D data due to its unique characteristics.   

\begin{table}[!h]
\caption{Image-level anomaly detection performance in terms of I-AUROC.}\label{tab:performance image-auroc}

\resizebox{\textwidth}{!}{%
\begin{tabular}{cl|cccccccccc|c}
\hline
 & Method & Bagel & \makecell{Cable\\Gland} & Carrot & Cookie & Dowel & Foam & Peach & Potato & Rope & Tire & \textbf{Mean} \\ 
\hline
\multirow{5}{*}{RGB } 
 & BTF~\cite{horwitz2023back} & 0.876 & 0.880 & 0.791 & 0.682 & 0.912 & 0.701 & 0.695 & 0.618 & 0.841 & 0.702 & 0.770 \\
 & AST~\cite{rudolph2023asymmetric} & \textbf{0.947} & 0.928 & 0.851 & \textbf{0.825} & \textbf{0.981} & \textbf{0.951} & 0.895 & 0.613 & \textbf{0.992} & 0.821 & \textbf{0.880} \\
 & M3DM~\cite{wang2023multimodal} & 0.944 & 0.918 & 0.896 & 0.749 & 0.959 & 0.767 & \textbf{0.919} & 0.648 & 0.938 & 0.767 & 0.850 \\
 & Shape-guided~\cite{chu2023shape} & 0.911 & \textbf{0.936} & 0.883 & 0.662 & 0.974 & 0.772 & 0.785 & 0.641 & 0.884 & 0.706 & 0.815 \\
 & \textbf{MDSS} & 0.915 & 0.894 & \textbf{0.907} & 0.780 & 0.963 & 0.793 & 0.869 & \textbf{0.743} & 0.953 & \textbf{0.856} & 0.867 \\ 
\hline
\multirow{5}{*}{3D} 
& BTF~\cite{horwitz2023back} & 0.825 & 0.551 & 0.952 & 0.797 & 0.883 & 0.582 & 0.758 & 0.889 & 0.929 & 0.653 & 0.782 \\
 & AST~\cite{rudolph2023asymmetric} & 0.881 & 0.576 & 0.965 & 0.957 & 0.679 & 0.797 & 0.990 & 0.915 & 0.956 & 0.611 & 0.833 \\
 & M3DM~\cite{wang2023multimodal} & 0.941 & 0.651 & 0.965 & 0.969 & 0.905 & 0.760 & 0.880 & 0.974 & 0.926 & 0.765 & 0.874 \\
 & Shape-guided~\cite{chu2023shape} & \textbf{0.983} & 0.682 & \textbf{0.978} & \textbf{0.998} & \textbf{0.960} & 0.737 & \textbf{0.993} & 0.979 & \textbf{0.966} & 0.871 & \textbf{0.916} \\
 & \textbf{MDSS} & 0.969 & \textbf{0.691} & 0.959 & 0.906 & 0.849 & \textbf{0.865} & 0.966 & \textbf{0.989} & 0.898 & \textbf{0.926} & 0.902 \\ 
\hline
\multirow{6}{*}{RGB+3D} 
& BTF~\cite{horwitz2023back} & 0.918 & 0.748 & 0.967 & 0.883 & 0.932 & 0.582 & 0.896 & 0.912 & 0.921 & 0.886 & 0.865\\
 & AST~\cite{rudolph2023asymmetric} & 0.983 & 0.873 & 0.976 & 0.971 & 0.932 & 0.885 & 0.974 & \textbf{0.981} & \textbf{1.000} & 0.797 & 0.937 \\
 & M3DM~\cite{wang2023multimodal} & \textbf{0.994} & 0.909 & 0.972 & 0.976 & 0.960 & 0.942 & 0.973 & 0.899 & 0.972 & 0.850 & 0.945 \\
 & Shape-guided~\cite{chu2023shape} & 0.986 & 0.894 & 0.983 & 0.991 & 0.976 & 0.857 & \textbf{0.990} & 0.965 & 0.960 & 0.869 & 0.947 \\
 & CFM\tablefootnote{Recently accepted in CVPR 2024 (\url{https://cvpr.thecvf.com/Conferences/2024/AcceptedPapers})}~\cite{costanzino2023multimodal} & \textbf{0.994} & 0.888 & \textbf{0.984} & \textbf{0.993} & \textbf{0.980} & 0.888 & 0.941 & 0.943 & 0.980 & \textbf{0.953} & 0.954 \\
 & \textbf{MDSS} & 0.983 & \textbf{0.911} & \textbf{0.984} & 0.927 & 0.955 & \textbf{0.962} & 0.973 & 0.978 & 0.962 & 0.930 & \textbf{0.956} \\ 
\hline
\end{tabular}}%
\end{table}

\begin{table}[!h]
\caption{Pixel-level anomaly performance in terms of AUPRO.}\label{tab:performance aupro}
\resizebox{\textwidth}{!}{%
\begin{tabular}{cl|cccccccccc|c}
\hline
& Method & Bagel & \makecell{Cable \\ Gland} & Carrot & Cookie & Dowel & Foam & Peach & Potato & Rope & Tire & \textbf{Mean}\\
\hline
\multirow{4}{*}{RGB} 
 & BTF~\cite{horwitz2023back} & 0.901 & 0.949 & 0.928 & 0.877 & 0.892 & 0.563 & 0.904 & 0.932 & 0.908 & 0.906 & 0.876\\
 & M3DM~\cite{wang2023multimodal} & \textbf{0.952} & \textbf{0.972} & 0.973 & 0.891 & 0.932 & \textbf{0.843} & \textbf{0.970} & 0.956 & \textbf{0.968} & 0.966 & \textbf{0.942} \\
 & Shape-guided~\cite{chu2023shape} & 0.946 & \textbf{0.972} & 0.960 & \textbf{0.914} & \textbf{0.958} & 0.776 & 0.937 & 0.949 & 0.956 & 0.957 & 0.933 \\
 & \textbf{MDSS} & 0.917 & 0.967 & \textbf{0.975} & 0.873 & 0.951 & 0.808 & 0.935 & \textbf{0.969} & 0.949 & \textbf{0.977} & 0.932 \\ 
\hline
\multirow{4}{*}{3D} 
 & BTF~\cite{horwitz2023back} & 0.973 & \textbf{0.879} & \textbf{0.982} & 0.906 & 0.892 & 0.735 & 0.977 & 0.982 & \textbf{0.956} & 0.961 & 0.924 \\
 & M3DM~\cite{wang2023multimodal} & 0.943 & 0.818 & 0.977 & 0.882 & 0.881 & 0.743 & 0.958 & 0.974 & 0.950 & 0.929 & 0.906 \\
 & Shape-guided~\cite{chu2023shape} & \textbf{0.974} & 0.871 & 0.981 & \textbf{0.924} & \textbf{0.898} & 0.773 & 0.978 & \textbf{0.983} & 0.955 & \textbf{0.969} & \textbf{0.931} \\
 & \textbf{MDSS} & 0.973 & 0.818 & 0.979 & 0.911 & 0.874 & \textbf{0.801} & \textbf{0.982} & \textbf{0.983} & 0.949 & 0.960 & 0.923 \\ 
 \hline
\multirow{5}{*}{RGB+3D} 
& BTF~\cite{horwitz2023back} & 0.976 & 0.969 & 0.979 & \textbf{0.973} & 0.933 & 0.888 & 0.975 & 0.981 & 0.950 & 0.971& 0.959 \\
& M3DM~\cite{wang2023multimodal} & 0.970 & 0.971 & 0.979 & 0.950 & 0.941 & 0.932 & 0.977 & 0.971 & 0.971 & 0.975 & 0.964 \\
 & Shape-guided~\cite{chu2023shape} & \textbf{0.981} & \textbf{0.973} & \textbf{0.982} & 0.971 & \textbf{0.962} & \textbf{0.978} & 0.981 & 0.983 & 0.974 & 0.975 & \textbf{0.976} \\
 & CFM~\cite{costanzino2023multimodal} & 0.979 & 0.972 & \textbf{0.982} & 0.945 & 0.950 & 0.968 & 0.980 & 0.982 & \textbf{0.975} & \textbf{0.981} & 0.971 \\
 & \textbf{MDSS} & 0.979 & 0.968 & 0.981 & 0.949 & 0.958 & 0.969 & \textbf{0.982} & \textbf{0.983} & 0.970 & 0.978 & 0.972 \\
 \hline
\end{tabular}}%
\end{table}

As shown in Table~\ref{tab:performance image-auroc}, our method MDSS obtains the state-of-the-art performance in terms of average I-AUROC over all categories for multimodal anomaly detection. To be specific, MDSS achieves the mean I-AUROC with 0.956, outperforming the best student-teacher network method AST and the memory bank based method Shape-guided by 1.9\% and 0.9\% respectively. Compared with the latest method CFM also without memory bank, MDSS still demonstrates superior performance.

In addition, when only using RGB images or 3D point clouds for anomaly detection, MDSS both ranks second among the employed methods. Particularly, MDSS only performs worse than AST for RGB images and Shape-guided for 3D point clouds. This indicates that for image-level anomaly detection in terms of I-AUROC, MDSS is effective and stable whether RGB images or 3D point clouds or both are used.

From Table~\ref{tab:performance aupro}, it could be observed that MDSS obtains the mean AUPRO with 0.972 for anomaly segmentation in multimodal setting, sightly worse than Shape-guided but better than other three methods (including the latest method CFM). Moreover, BTF and M3DM perform unstable for different modals. For example, M3DM ranks first for RGB images while ranks last for 3D point clouds. This may explain why BTF and M3DM perform worse than MDSS when combining RGB images and 3D point clouds for anomaly segmentation.

\begin{figure}[!h] 
\includegraphics[width=0.95\textwidth]{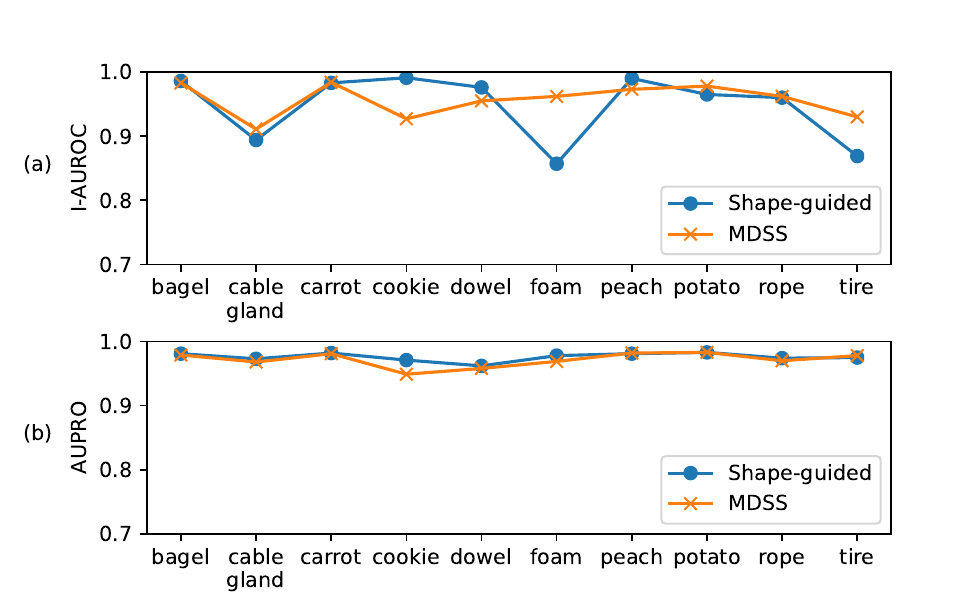}
\centering
\caption{Comparison of anomaly detection performance between MDSS and Shape-guided across all categories.} \label{fig:Comparison}
\end{figure}

In summary, MDSS and Shape-guided are the two best methods respectively for multimodal anomaly detection and segmentation. For the purpose of  comparing these two methods more comprehensively, Fig.~\ref{fig:Comparison} provides the detailed comparison for each category. 

From Fig.~\ref{fig:Comparison}, it could be observed that for anomaly segmentation, MDSS and Shape-guided show relatively close performance in terms of AUROC for different categories, which means that these two methods are both stable and effective for anomaly segmentation. However, for anomaly detection, we can observe the performance fluctuation for Shape-guided in terms of I-AUROC, indicating the instability of Shape-guided for anomaly detection.  

To conclude, MDSS is comparable but more stable than the SOTA memory bank based method Shape-guided, and furthermore performs better than other baseline methods in multimodal anomaly detection.

\subsection{Ablation Study} 

\begin{figure}[!h] 
\includegraphics[width=0.75\textwidth]{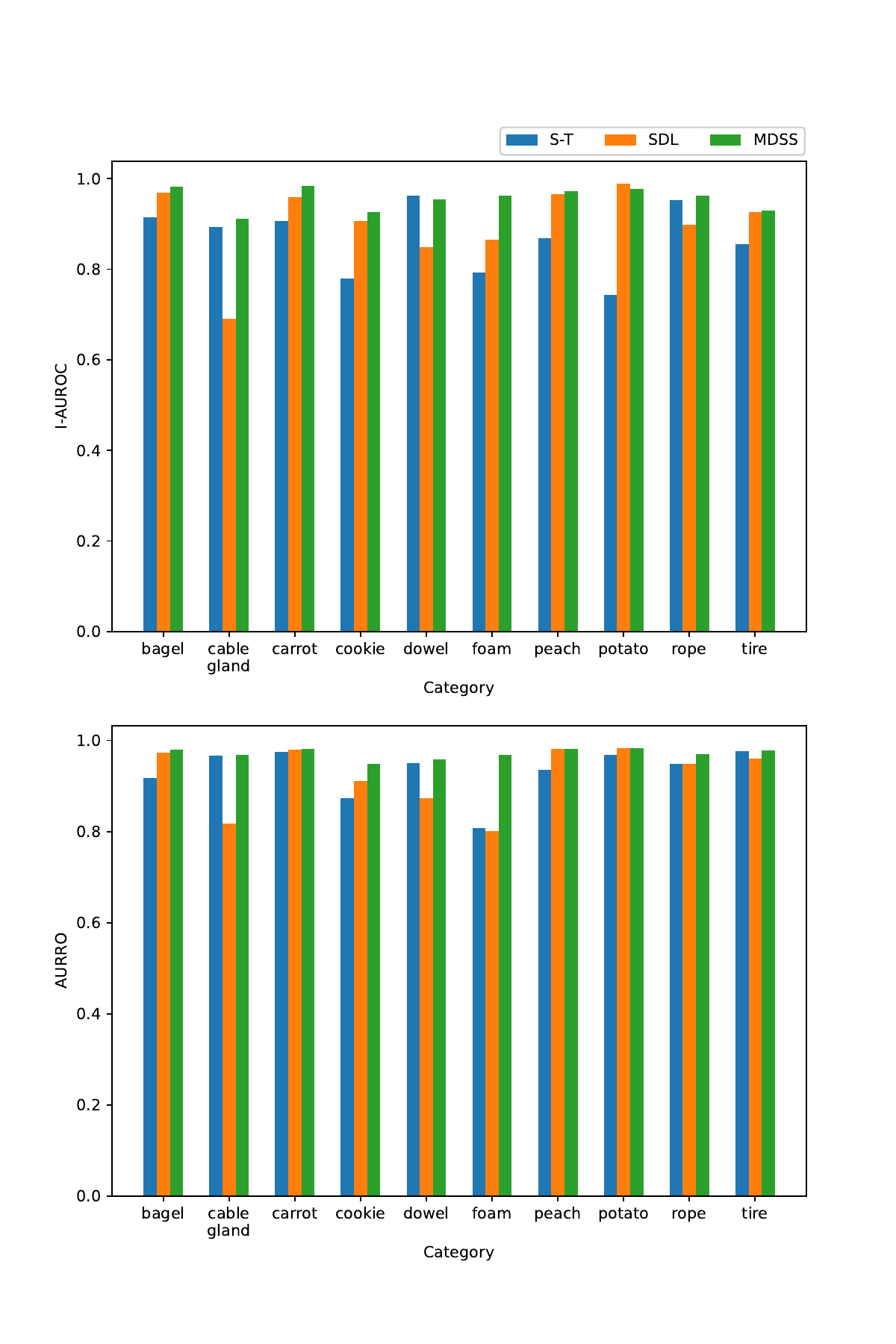}
\centering
\caption{Ablation study for MDSS. Anomaly detection performance of MDSS in S-T only, SDL only, and combined settings.}
\label{fig:ablation}
\end{figure}

We conduct an ablation study to demonstrate the effectiveness of combining the student-teacher network and signed distance learning.
Fig.~\ref{fig:ablation} provides the detailed comparison of MDSS, student-teacher (S-T) network and signed distance learning (SDL) for different categories in terms of I-AUROC and AUPRO.

From Fig.~\ref{fig:ablation}, it could be observed that MDSS usually  performs better than S-T and SDL for most categories. Specifically, MDSS obtains the best I-AUROC for eight categories ( except dowel and potato) and achieves the best AUPRO for all categories. As is known to all, some anomalies may manifest in terms of color while others may manifest in terms of 3D geometry, combining the two modality information for detecting anomaly will be more effective. 

In conclusion, MDSS activates the interaction between the two modality information and usually performs better than the two individual modules in MDSS, namely student-teacher network and signed distance learning.  

\section{Conclusion}
In present study, a novel memoryless multimodal anomaly detection method MDSS is proposed, which includes three different modules, namely student-teacher network, signed distance learning and score map alignment. Specifically, student-teacher network aims to learn RGB images and masks generated from 3D point clouds to obtain the RGB anomaly score map. In signed distance learning, we employ the signed distance function to reconstruct surface from normal point clouds and adopt the distance from the point to the surface to generate corresponding 3D anomaly score map. Finally, a statistical approach is employed to align the RGB anomaly score map and 3D anomaly score map, and the aligned score map which combines the anomaly information from both RGB and 3D is used for anomaly detection. The experimental results with popular MVTec 3D-AD dataset demonstrate that MDSS is comparable but more stable than the SOTA memory bank based method Shape-guided, and furthermore performs better than other baseline methods. Our method demonstrates that for multimodal anomaly detection, performance is still on the rise without the usage of memory banks, making it more suitable for some real-world application scenarios.

\bibliographystyle{splncs04}
\bibliography{mybibliography}
\end{document}